\documentclass[conference]{IEEEtran}
\usepackage{amsmath,graphicx}

\usepackage{url}
\usepackage{multirow}

\def\BibTeX{{\rm B\kern-.05em{\sc i\kern-.025em b}\kern-.08em
		T\kern-.1667em\lower.7ex\hbox{E}\kern-.125emX}}

\title{Sustainability using Renewable Electricity (SuRE) towards NetZero Emissions}

\author{\IEEEauthorblockN{Jinu Jayan, Saurabh Pashine, Pallavi Gawade, Bhushan Jagyasi, Sreedhar Seetharam, Gopali Contractor, \\
Rajesh kumar Palani, Harshit Sampgaon, Sandeep Vaity, Tamal Bhattacharyya, Rengaraj Ramasubbu}
\IEEEauthorblockA{Artificial Intelligence Practice, Accenture, India \\ 
\{jinu.jayan, saurabh.pashine, pallavi.s.gawade, bhushan.jagyasi, sreedhar.p.seetharam, gopali.contractor, \\ 
	rajesh.kumar.palani, harshit.sampgaon, sandeep.vaity, 
	tamal.bhattacharyya, rengaraj.ramasubbu\}@accenture.com}
}

\begin{document}
\maketitle
\begin{abstract}
	Demand for energy has increased significantly across the globe due to increase in population and economic growth. Growth in energy demand poses serious threat to the environment since majority of the energy sources are non-renewable and based on fossil fuels, which leads to emission of harmful greenhouse gases. Organizations across the world are facing challenges in transitioning from fossil fuels-based sources to greener sources to reduce their carbon footprint. 
	\\ 
	As a step towards achieving Net-Zero emission target, we present a scalable AI based solution that can be used by organizations to increase their overall renewable electricity share in total energy consumption. 
	Our solution provides facilities with accurate energy demand forecast, recommendation for procurement of renewable electricity to optimize cost  and carbon offset recommendations to compensate for Greenhouse Gas (GHG) emissions. 
	This solution has been used in production for more than a year for four facilities and has increased their renewable electricity share significantly. 
\end{abstract} 
\section{Introduction} 
Many countries have pledged to reduce overall carbon emission and achieve Net-Zero emissions by 2050 \cite{IEA_NetZero}. 
There are different sectors - Energy, Agriculture, Forestry and land use, Waste and Industry that contribute to the global green house gas emissions.  
Among these, Energy contributes to 73.2\% of green house gas emissions \cite{OurWorldindata}. 
To achieve Net-Zero emissions by 2050, major portion of this reduction is hence attributed to come by increasing the renewable energy share in the overall consumption. 

Ever since the Paris agreement of 2016 \cite{ParisAgreement}, countries across the world are actively finding ways to reduce global warming to less than 2$^{\circ}$C compared to the pre-industrial levels. There are commitments made by countries, via Nationally Determined Contributions (NDCs), listing their target contribution in each sector. 
These mandates flow down to all businesses operating within a country to make them sustainable. 
Many countries have introduced a range of carbon emissions reduction policies including mandatory carbon emission capacity, carbon cap, carbon emission tax, cap-and-trade, carbon offset, and joint implementation to curb the total amount of carbon emissions \cite{UNFCCC_NDC}.

\begin{figure}[h]
	\centering
	\includegraphics[width=.95\columnwidth]{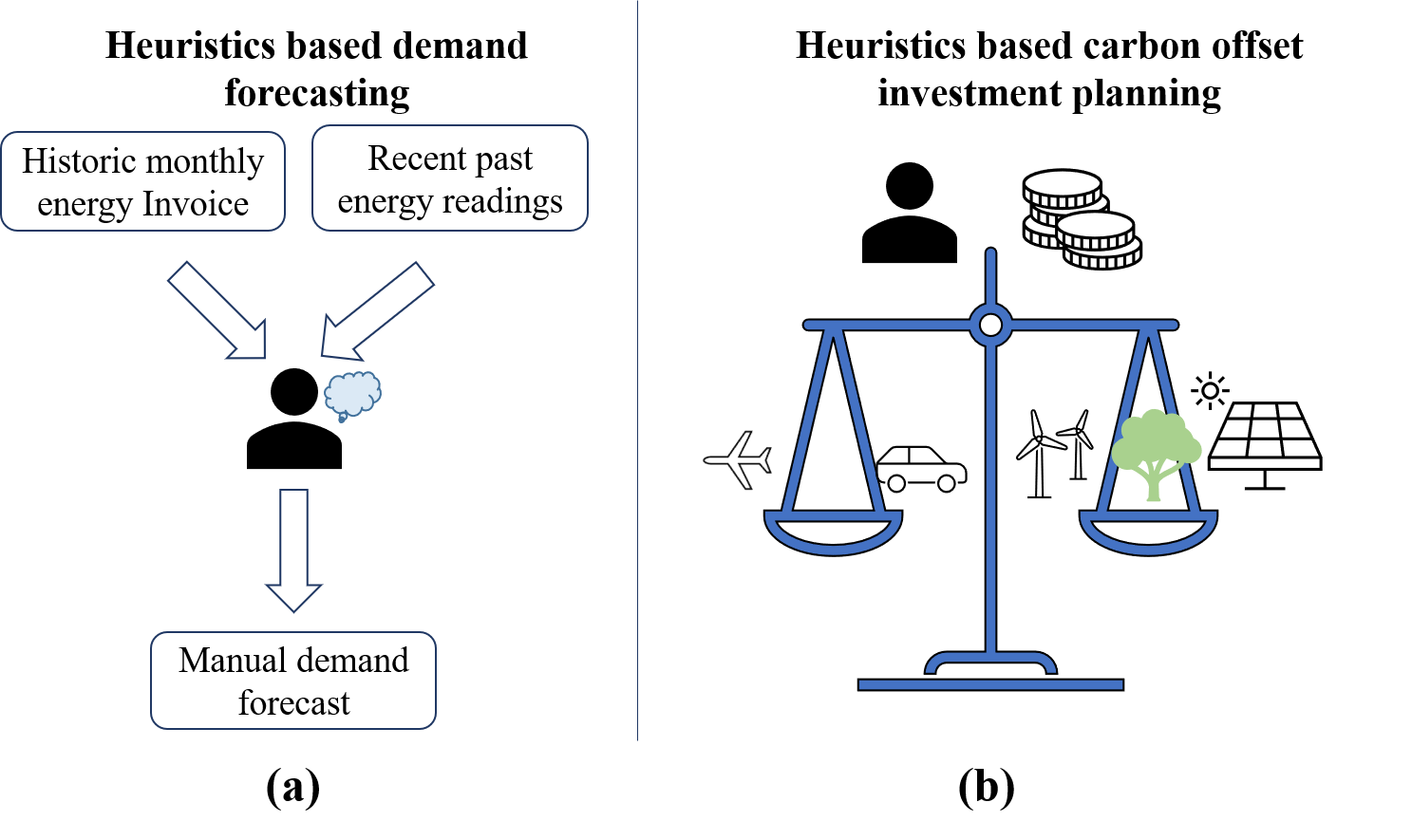}
	\caption{(a) Existing energy management process for commercial buildings (b) Carbon Offset - A difficult balancing act}
	\label{ExistingProcess}
\end{figure}

In this paper, we focus on providing solution for commercial buildings to achieve Net-Zero emissions target.   
The buildings are typically powered by one or multiple electricity sources, from different conventional and green energy suppliers. 
We consider a scenario where commercial buildings are powered by both conventional and green energy sources procured from external power generator vendors. As per the vendor contract agreement, the building administrator is expected to provide monthly renewable electricity demand at the start of the month. 
As depicted in Figure \ref{ExistingProcess}, there was no holistic end-to-end energy management solution to manage energy demand, reduce carbon footprint and achieve their goals of Net-Zero emissions. The demand estimation was performed manually and long term energy demand was not visible for carbon offset planning.
Hence we required an accurate forecast to maximize the utilization of green energy consumption which in turn can reduce the overall carbon emission. In geographies with no scope of increasing renewable energy through such advance estimations, we can only use the offset planning feature of the solution.  

This solution is currently been deployed for the organization's Indian business, where renewable energy is supplied by the green power vendors through the electricity grid maintained by the state electricity provider. If the actual consumption is more than the ask, the additional units are covered from the conventional energy. And if the actual consumption is less than the green power asked, the monthly invoice will be billed based on the total units asked. Hence to maximize the consumption of renewable electricity and to minimize the cost we need an accurate forecast of an overall energy demand for the month. 

The problem of forecasting electricity demand in a large organization with thousands of employees in a building is challenging. This is due to uncertainties introduced by weather, building occupancy, grid outages and measurement noises. 

The key contributions of this paper are: 
\begin{itemize}
\item Sustainability using Renewable Electricity (SuRE) application for accurate energy demand forecasting and clean energy recommendation system
\item AI models for daily occupancy forecasting in commercial buildings 
\item Efficient approach for carbon offset recommendation towards achieving NetZero emissions target
\item Insights from pilot deployment of SuRE end-to-end AI based energy management system in four large commercial facilities for a year
\end{itemize}

In the next section we present the related work, followed by proposed SuRE Application in sections \ref{app_layout} - \ref{ai_models}. The impact of the deployed SuRE application is then covered in section \ref{app_deployment} and \ref{results}. Finally, we conclude the paper in section \ref{conclusion}.  

\section{Literature Review}
With increasing need for sustainable economic growth and also to mitigate the effects of climate change, there is a large focus on innovative ways to achieve NetZero emission targets. Since global electricity contributes to 40\% of these emissions \cite{Hindawi_CO2E}, there has been a large amount of research dedicated to improve ways of generating and consuming electricity so as to reduce its carbon footprint. 
This will require accurate estimation of demand to provide optimal plan for equipment operations, budgeting, and energy procurement.

For a given consumer, short term demand forecasting will help in operating hourly load optimally and medium to long term forecasts for budgeting and energy planning. Uncertainties in the demand side consumption is described in \cite{Shi2018}. 
Forecasting techniques can fall broadly under two categories of classical statistical techniques and Artificial Neural Network (ANN) approach \cite{Ali2015}. Classical techniques include multiple linear regression \cite{Amral2007}, exponential smoothing \cite{Taylor2008}, auto regressive integrated moving average (ARIMA) \cite{Hwang1995} and Kalman ﬁltering \cite{Soliman2004}, \cite{Feng2007}. ANNs are also shown to be performing well, especially when there is a need to model non linearity \cite{Kandil2006}, \cite{Azadeh2006}. 

Medium term forecasts are then used as input to an optimization model to determine the carbon offset plan for the building. There exists many multilateral and governmental reports on how to achieve Net-zero by 2050 \cite{IEA} and to limit global temperature rise to 2$^{\circ}$C, but not many studies and recommendations are available to achieve the target at per building/organization level. Also, not much research and literature are available on bridging demand forecasting with long term carbon offset planning. Here we propose an application of machine learning models for forecasting and constraint optimization to generate an optimal plan to achieve NetZero emissions.

\section{Proposed Sustainability using Renewable Electricity (SuRE) application}
\label{app_layout}
\begin{figure*}[!h]
	\centering
	\includegraphics[width=1.55\columnwidth]{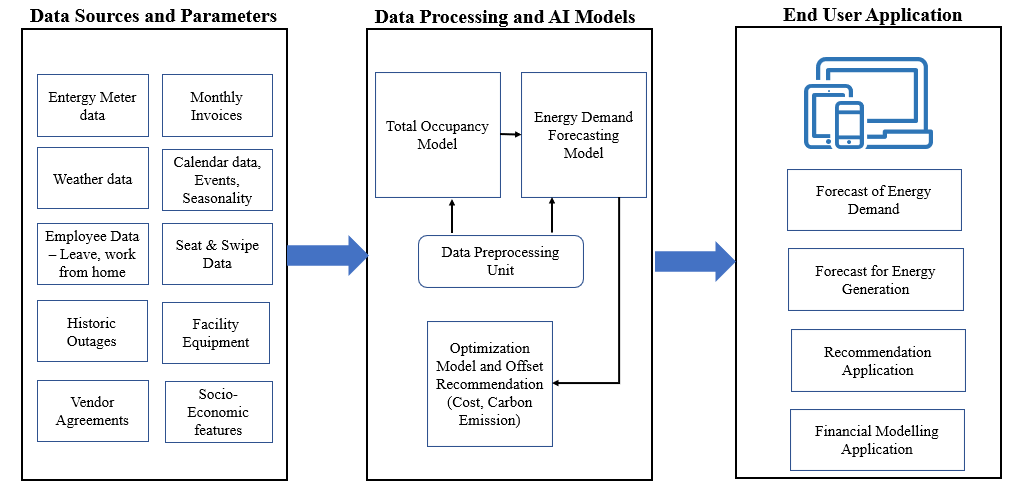} 
	\caption{Proposed architecture diagram of SuRE (Sustainability using Renewable Electricity) application}.
	\label{architecture}
\end{figure*}

In this section, we provide details of our proposed Sustainability using Renewable Electricity (\emph{SuRE}) application. 
The architecture diagram of \emph{SuRE} is shown in Figure \ref{architecture}.

\begin{enumerate}
\item 
Various parameters required for training and inference of the machine learning models are depicted under the \emph{Data Sources and Parameters} in Figure \ref{architecture}. The details of these parameters are covered in section \ref{data_section}. 
 
\item \emph{Data Processing and AI Models} layer houses the automated data pre-processing along with several models. 
Different components of SuRE application include - Occupancy Model, Energy Demand Forecasting Model, Optimization Model and Recommendation Engine. These components are described in section \ref{ai_models}.

\item In section \ref{app_deployment}, we present an end user application which provides short-term \& medium-term forecast and analysis of accuracy for historical electricity consumption for the facility manager. 
It further provides recommendation on offset plans for minimizing carbon emissions.
\end{enumerate}
\section{Data Sources and Parameters}
\label{data_section}
The organization selected for the study is a consulting major with large number of employees across multiple locations globally. 
This organization is actively working towards achieving 100\% Renewable Electricity (RE100) target by 2023. 
This requires increase in green power share and investments in offset mechanisms to achieve Net-Zero emissions. 

Data from multiple sources was collected from four different facilities in one of the operation centers in India. 
Extensive discussions were held with the business teams to understand the various parameters that can impact electricity consumption. 
Some of the parameters identified were - day of the week, daily employee attendance, seasonal factors like weather data, local holidays and events in the facility.

In this case, there was no centralized repository to access these varied data sources, as different parameters were captured by different applications and some data points were residing with the facilities. 
A unified data repository, by aggregating different data sources was hence created resulting in historical data for a duration of two years from four different facilities. This resulted in a rich dataset to conduct this research. 

\begin{figure}[h]
	\centering
	\includegraphics*[width=1\columnwidth]{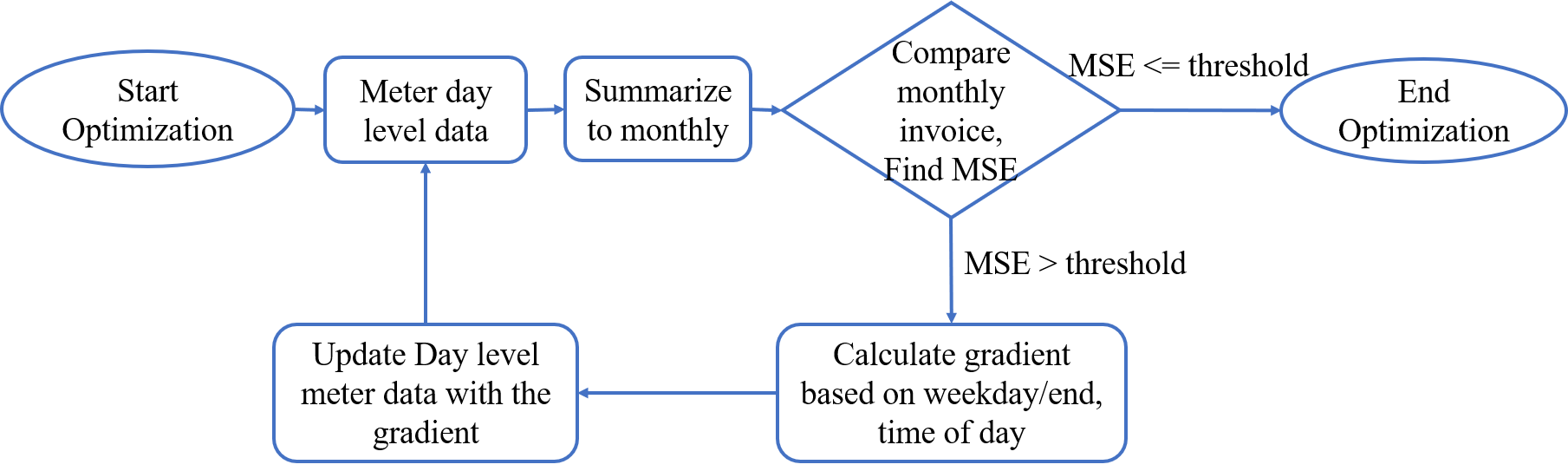} 
	\caption{Data Imputation Algorithm}
	\label{DataCuration}
\end{figure}

The energy meter data was collected from the smart metering systems which captures hourly power consumption in kWh. This data had anomalies like missing values which was caused due to issues in the way the data was captured and stored. Some of the issues were due to smart meters getting frozen with no data being tracked, replacement of smart meters, connectivity issues between the meter and the back end database systems. These issues created big gaps in the time series data. To overcome this problem and to create a consistent and continuous time series data, we developed a new imputation strategy as shown in Figure \ref {DataCuration}. 
Missing data was imputed by taking consumption data obtained from the monthly invoices as reference and using a gradient descent optimization we iteratively updated the values. We initialized the missing values using historic values obtained based on day of the week and month of the year and then adapted these values till monthly mean squared error (MSE) (optimization cost) was reduced below a configured threshold (1\% used in this case). Monthly MSE is computed using estimated monthly consumption (sum of estimated daily consumption for a month) compared with the consumption extracted from the actual invoices.

To estimate the number of people visiting (daily occupancy) in the building, we collected other important parameters like employee swipe data, scheduled events in the facility and local holidays. 
Further, weather parameters for the day like average temperature, average precipitations, average humidity and pressure were also obtained through external weather services. 
The data spans the duration of before and after the start of Covid-19 pandemic. There were large scale changes in consumption trends due to nationally mandated lock downs. To make full use of the available two years of data, we have added a feature to identify if the data point belonged to either lockdown or non-lockdown period.

\section{Data Processing and AI Models}
\label{ai_models}
The Data processing and AI models layer as shown in Figure \ref{architecture} consists of a Data pre-processing unit, Total Occupancy model, Energy Demand Forecasting model and Optimization model. Data from the unified data repository is read by the consumer services within the data pre-processing unit. Data is then passed to the data pre-processing engine which consists of data validation, correction and imputation which are performed in sequence. Different datasets are then merged into a single dataset. As depicted in Figure 2, there exists three separate models and the data flows serially though these models. 
These models include: (a) Total Occupancy model (b) Energy Demand forecasting model and (c) Optimized carbon offset planning model. Each model is dependent on the output coming from the upstream model. Once pre-processing of the data is completed, it is first utilized for prediction of daily occupancy using Total Occupancy Model.
The predicted occupancy along with other parameters like weather, seasonality, events calendar is further consumed by the Energy Demand Forecasting Model to forecast daily and aggregated monthly electricity demand for the entire facility. 
Finally, the carbon offset planning model uses the medium-term demand forecasts and other context specific inputs like local tax incentives, generation capacities of the organization to estimate the net emissions liability. For this emissions liability, the model provides recommendation on green energy procurement, generation and other alternate energy offset plans to achieve Net-Zero carbon emissions.

\subsection{Total Occupancy Forecasting Model}
\begin{figure}[h]
	\centering
	\includegraphics[width=1\columnwidth]{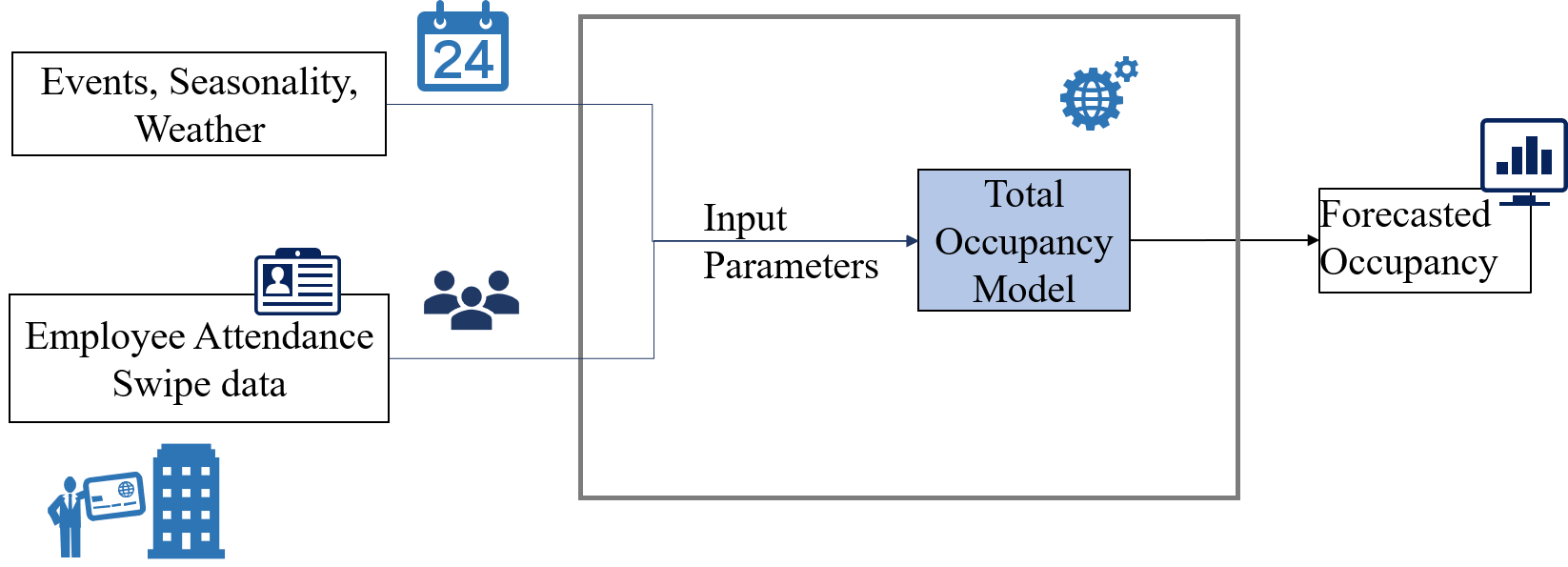} 
	\caption{Occupancy Model}
	\label{OccupancyModel}
\end{figure}

Daily power consumption in any building have a strong dependency on footfalls or occupancy in the building. 
Hence, an occupancy model has been developed for forecasting daily occupancy in building which can in-turn be used as a feature to the daily energy demand. 
As shown in the Figure \ref{OccupancyModel}, an occupancy model is developed using features like employee swipe data, seasonality, visitor counts, holidays and event schedule. 
Ensemble models like Random Forest, Boosted Trees, Light Gradient Boosting Machine are used in the ML engine of occupancy model. Automated test and evaluations as detailed in the subsection \ref{model_details} are used to estimate the short-term and medium-term occupancy within the building.

\subsection{Energy Demand Forecasting Model}
\begin{figure}[h]
 	\centering
 	\includegraphics[width=1\columnwidth]{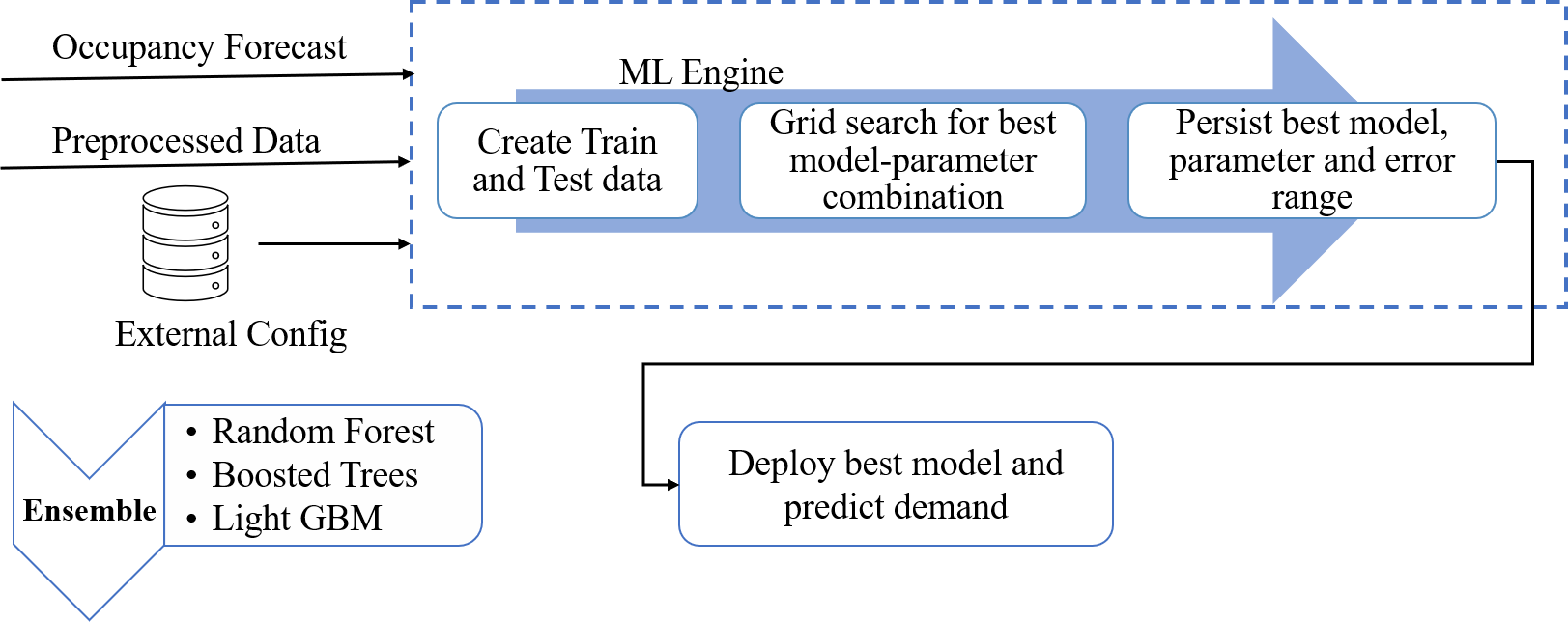}
 	\caption{Demand Forecast Model}
 	\label{DemandForecastModel}
 \end{figure}

 To forecast short-term and medium-term energy demand (Figure \ref{DemandForecastModel}), we make use of the occupancy estimates generated by the occupancy model along with  weather, holiday, recent consumption trends and day of the week. Recent consumption trends, is an engineered feature which captures the average consumption in the last week. As in occupancy model, ensemble models like Random Forest, Boosted Trees, Light Gradient Boosting Machine are used in the ML engine to arrive at the best model. The details of model evaluation are similar to the process followed for occupancy model and are described in the subsection \ref{model_details}.
 The best model will then be used to conduct the inference for short-term and medium-term energy demand estimations for future dates. 
 
\begin{table}
    \centering
    \caption{Model Specifications}
    \begin{tabular}{|p{0.18\linewidth}|p{0.24\linewidth}|p{0.42\linewidth}|}
   
    \hline
    Algorithm & Parameter  & Range \\
    \hline
    \multirow{3}{*}{\parbox{1.5cm}{Random Forest}} & max features & \{0.5, 0.6\} \\ 
    \cline{2-3}
    & n estimators & \{10, 16, 32, 40, 50\} \\
    \cline{2-3}
    & random state & \{42\}\\
    \hline
    \multirow{3}{*}{\parbox{1.5cm}{Boosted Trees}} & learning rate  & \{0.01, 0.05, 0.1, 0.15, 0.2, 0.25, 0.3\}\\
    \cline{2-3}
    & n estimators & \{25, 50, 75, 100\} \\
    \cline{2-3}
    & random state & \{42\}\\
    \hline
    \multirow{3}{*}{\parbox{1.5cm}{Light GBM}} & learning rate & \{0.01, 0.05, 0.1, 0.15, 0.2\}\\
    \cline{2-3}
    & n estimators & \{10, 16, 32, 40, 50\} \\
    \cline{2-3}
    & random state & \{42\}\\
    \hline
    \end{tabular}
    
    \label{model_specs}
\end{table}

\subsection{Forecasting Model Details}
\label{model_details}
Daily samples of occupancy and energy consumption data was available starting from Jan 2019. New data as available at the end of every month is appended to the existing dataset. To account for the new data points being added to the time-series  dataset, a new model has been built every month. 

Various regression algorithms available within scikit-learn \cite{scikit-learn} and LightGBM \cite{lightgbm} python api are made use of for this purpose. As the data used is a time series data, traditional cross validation is replaced with cross-validation with sliding window. To validate the quality of the model and to ensure less variance during inference we conduct tests for various algorithms and parameters by predicting daily consumption units for previous two months for which the ground truth is available. For energy demand forecasting, predicted daily units are aggregated to generate monthly forecast which is compared with the actual consumption. 

We capture metrics like Mean Squared Error (MSE) and Adjusted R2. Mean Squared Error (MSE) is calculated using  $MSE = 1/n\sum_{i=1}^n (Y_{i}-\hat{Y}_{i})^2$  where $Y_{i}$ is the ground truth from monthly invoice and $\hat{Y}_{i}$ is the model predicted (summarized over a month) value for the $i^{th}$ predicted value and $n$ is the number of predictions.
We then compute $R^2_{adj} = 1 - \left[\frac{(1-R^2)(N-1)}{N-K-1}\right]$, 
where $R^2$ shows how well the data fit the model,
$N$ is the number of samples in the dataset, and
$K$ is the number of features in the dataset. 
$R^2_{adj}$  measures the percentage of variation that can be explained by the independent features. This metric will penalize features that do not contribute towards model fitness \cite{Dodge2008}.

Specifications of the various algorithms used in the solution are listed in Table \ref{model_specs}. The best models for Occupancy and Energy forecast are identified from within this combination of algorithm and parameters.
To illustrate, we take an example of a forecast scenario for the month of Aug-2020. To make estimations for the month we need to first identify a model which gives the best results for previous two months taken together. For a given algorithm and parameter combination we train two models. First model (M1) trained with data from Jan-2019 to May-2020, make predictions for June-2020. Second model (M2) trained with data from Jan-2019 to June-2020 use this model to make predictions for July-2020. Model-parameter combination resulting in the best average metrics are shown in Table \ref{model_performance}. 
\begin{table}
    \centering
    \caption{Performance of models for the month of Aug'20}
    \begin{tabular}{|p{0.18\linewidth}|p{0.3\linewidth}|p{0.1\linewidth}|p{0.09\linewidth}|}
    \hline
    Algorithm & Parameter  & MSE $10^{-6}$ & Adj R2 \\
    \hline
    Random Forest & max features: 0.9 \newline n estimators: 10 \newline random state: 42 & \vspace{0.225cm} 197.9  & \vspace{0.225cm} -0.42\\
    \hline
    Boosted Trees & learning rate: 0.15 \newline n estimators: 500 \newline random state : 42 &  \vspace{0.225cm}  131.9 & \vspace{0.225cm} -0.51\\
    \hline
    Light GBM  & learning rate: 0.2 \newline n estimators: 350 \newline random state : 42 & \vspace{0.225cm}  101.8  & \vspace{0.225cm} -0.22\\
    \hline
    \end{tabular}
    
    \label{model_performance}
\end{table}

As we move into the future, the new data will be appended to the existing dataset. Every month, new models are built and evaluated as the dataset is being continuously updated. To prevent uncontrolled growth of the dataset, only data not older than a certain number of months will only be stored, as very old data can have a very different characteristics as compared to current consumption patterns. This duration is a configurable parameter in the solution.

Overfitting is avoided by making the model robust against unseen future data. This generalization is achieved in two ways. Implicitly it is achieved within the ensemble models and explicitly via the moving window cross validations on recent historic months. Overfitting is also a function of model complexity. Model complexity is controlled via the number of estimators and number of features. These parameters are tuned to arrive at the most optimal type of models.

\subsection{Optimized carbon offset planning}
Carbon offset refers to the projects and programs to remove existing Greenhouse Gas (GHG) or preventing further emissions of GHGs into the atmosphere. To quantify the impact, the index considered is metric tonne of CO2 equivalent. This index is used to capture GHG reduction for CO2 or its equivalent of other gases. To achieve the required carbon offset, various projects are designed and executed. It includes projects like installing renewable energy plants like solar, wind, hydro, installation of energy efficient devices like light bulbs, cook stoves; plantations, forest rejuvenation, GHG removal from methane beds and industrial emissions. To ensure that these projects achieve the promised impact they are audited by accredited entities. 

To incentivize private participation in such projects there exists multilateral mechanisms through which entities get credits for their positive impact on the environment. Kyoto protocol \cite{KyotoProtocol} brought the Clean Development Mechanism (CDM) to validate and approve such projects. In turn the investing parties get credits in the form of clean energy certificates which can be traded and bought by entities who are planning to achieve Net-Zero emissions but are currently not having sufficient sources of renewable energy to runt their operations. 

 \begin{figure}[ht]
	\centering
	\includegraphics[width=1\columnwidth]{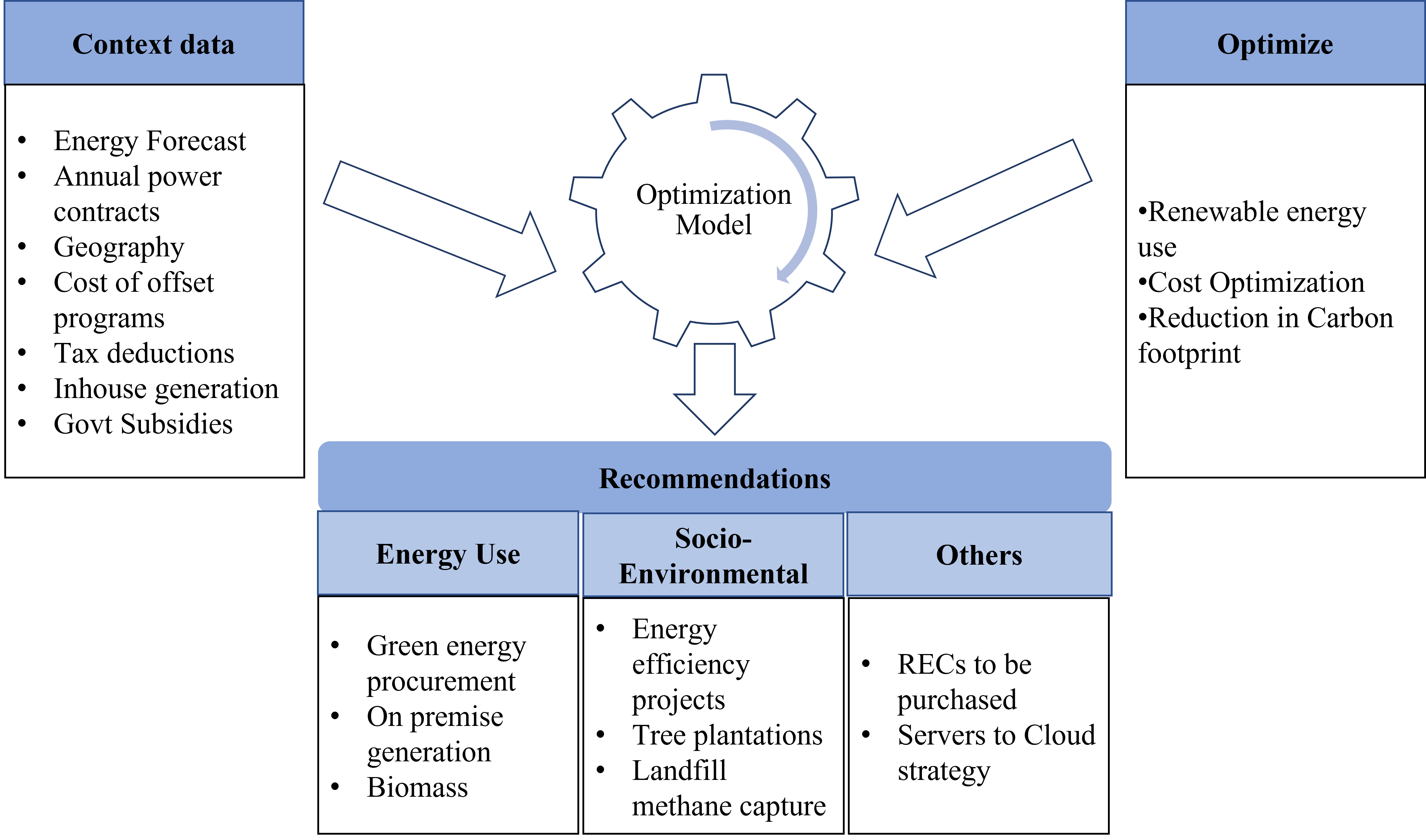} 
	\caption{Optimization Model}
	\label{Optimization}
\end{figure}

There is a limited literature on optimal planning for investments into such carbon projects, to be used by entities like a building or a regional unit of a business. As shown in the Figure \ref{Optimization} we use the medium-term forecast from the energy demand model along with contextual data like geography, local regulations, tax incentives, in-house generation capacities,  as some of the inputs. Along with context data we consider the offset capacity of each type of projects and their unit cost (cost of the project to remove one metric tonn of CO2 equivalent). These inputs are used to arrive at net emission liability for which an optimized recommendation will be generated listing out various offset projects like like - wind power, solar power, energy efficient devices and agroforestry to invest in. This optimization solution is implemented by making use of the constraint integer programming implementation provided by the Google OR tools \cite{ortools}.  

\subsubsection{Parameters:} 
Let $W_{pu}$, $S_{pu}$, $B_{pu}$, $E_{pu}$, and $A_{pu}$ be the per unit cost of offset projects - Wind power, Solar power, Biomass, Energy efficiency and Agro-forestry respectively.

\subsubsection{Constraints:}

For any project P, $P_{min}$ and $P_{max}$ defines the range of offset possible with that particular project. 
$P_{max} = J/P_{pu}$ and $P_{min} = J*X_P/P_{pu}$
where, $P_{pu}$ is the per unit cost for the offset project $P$, $J$ is the maximum possible total amount available for the investment in all offset projects and minimum $X_P\%$ of $J$ is expected to be invested in project $P$. 

Assuming there exists $W_{of}$, $S_{of}$, $B_{of}$, $E_{of}$, and $A_{of}$ as the optimal combination of the offsets for each of the projects. Then the investment constraint is defined as  $W_{of}*W_{pu} + S_{of}*S_{pu} +B_{of}*B_{pu}+E_{of}*E_{pu}+A_{of}*A_{pu} <= J$

\subsubsection{Objective:} 
Find $W_{of}$, $S_{of}$, $B_{of}$, $E_{of}$, and $A_{of}$ which satisfies all the above constraints and maximize the overall offset  i.e $Maximize (W_{of} + S_{of} + B_{of} + E_{of} + A_{of})$

Similar unique solution can also be determined if the user wants to achieve a specific amount of emissions offset. In this scenario we determine what is the most optimal allocation of investment towards each of the projects such that the offset goals are achieved with minimal amount. Hence a minimization objective.
 
\section{Application Deployment}
\label{app_deployment}
The solution was designed, developed, and deployed by the data science, architecture, application development and sustainability business teams. Azure cloud platform was selected to deploy the application. The application itself is deployed as a standalone Azure WebApp with integration to cloud storage. Organization facility teams use the application for their monthly forecast and renewable energy procurement. Medium term forecasting is used for energy budgeting and for carbon offset investment planning. An intuitive user interface is provided to interact with the backend ML and Optimization models. The application can be quickly deployed and used by any building management teams by only performing minimal configuration in an external configuration file.

\subsection{Application landing page}
The application home page provides the users an access to select specific building they are interested in. Within an organization the facility teams manage a number of buildings across various locations. Here they can invoke the models for individual towers for a specific month.

\subsection{Forecast and Historic Trend Analysis Interface}
\begin{figure}[h]
	\centering
	\includegraphics[width=1\columnwidth]{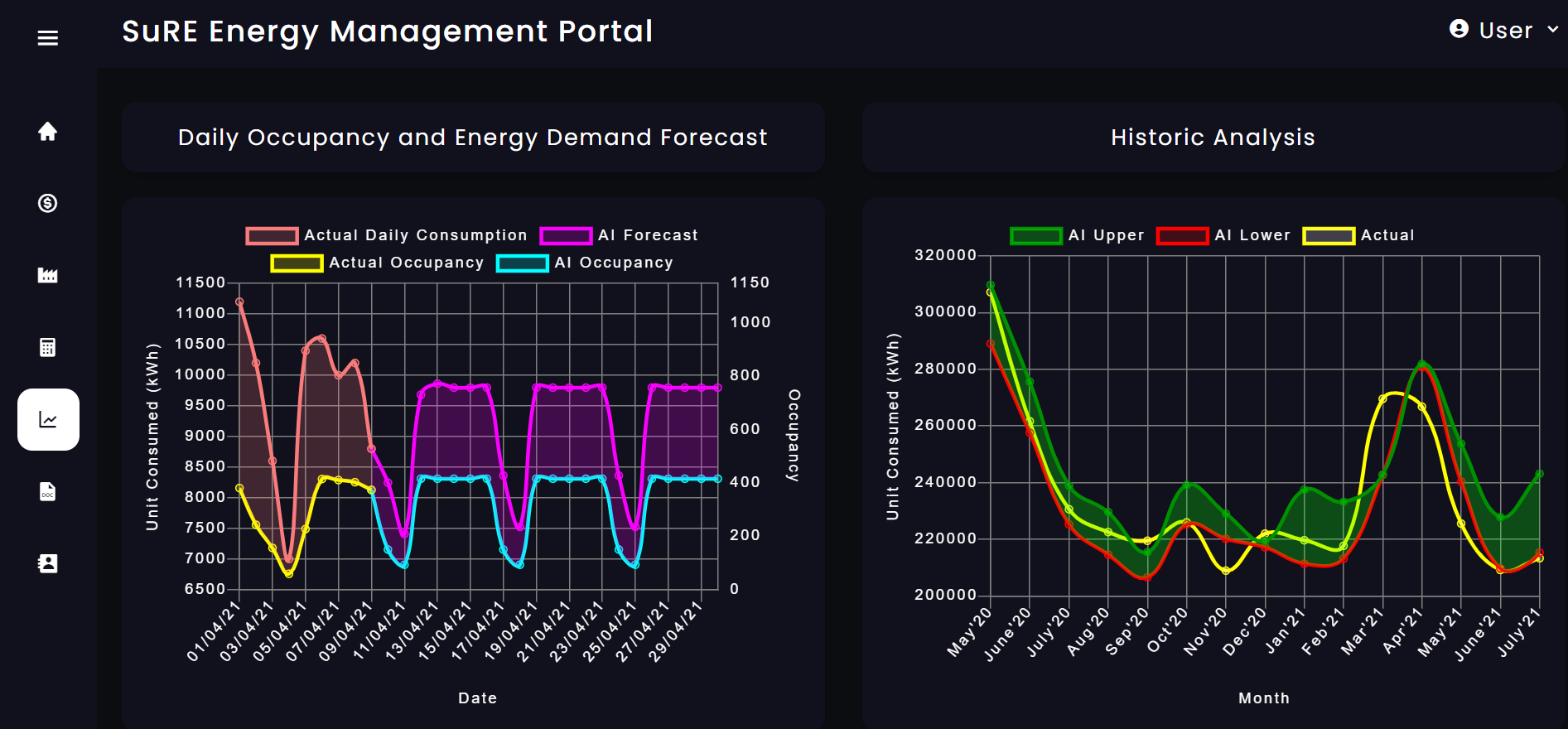} 
	\caption{(a) Forecasted Energy and Occupancy, (b) Historic Analysis of actual vs. forecasted demand}
	\label{ForecastingHistoricAnalysis}
\end{figure}

As shown in Figure \ref{ForecastingHistoricAnalysis}, this page provides forecasted daily energy demand and occupancy for a selected building and month. 
This page also provides a view of historic energy consumption with respect to forecast for the specific building and month. Multiple plots can be viewed simultaneously to understand how the models have fared as compared to the actual historic consumption.

\subsection{Financial Modeling Interface}
\begin{figure}[h]
	\centering
	\includegraphics[width=1\columnwidth]{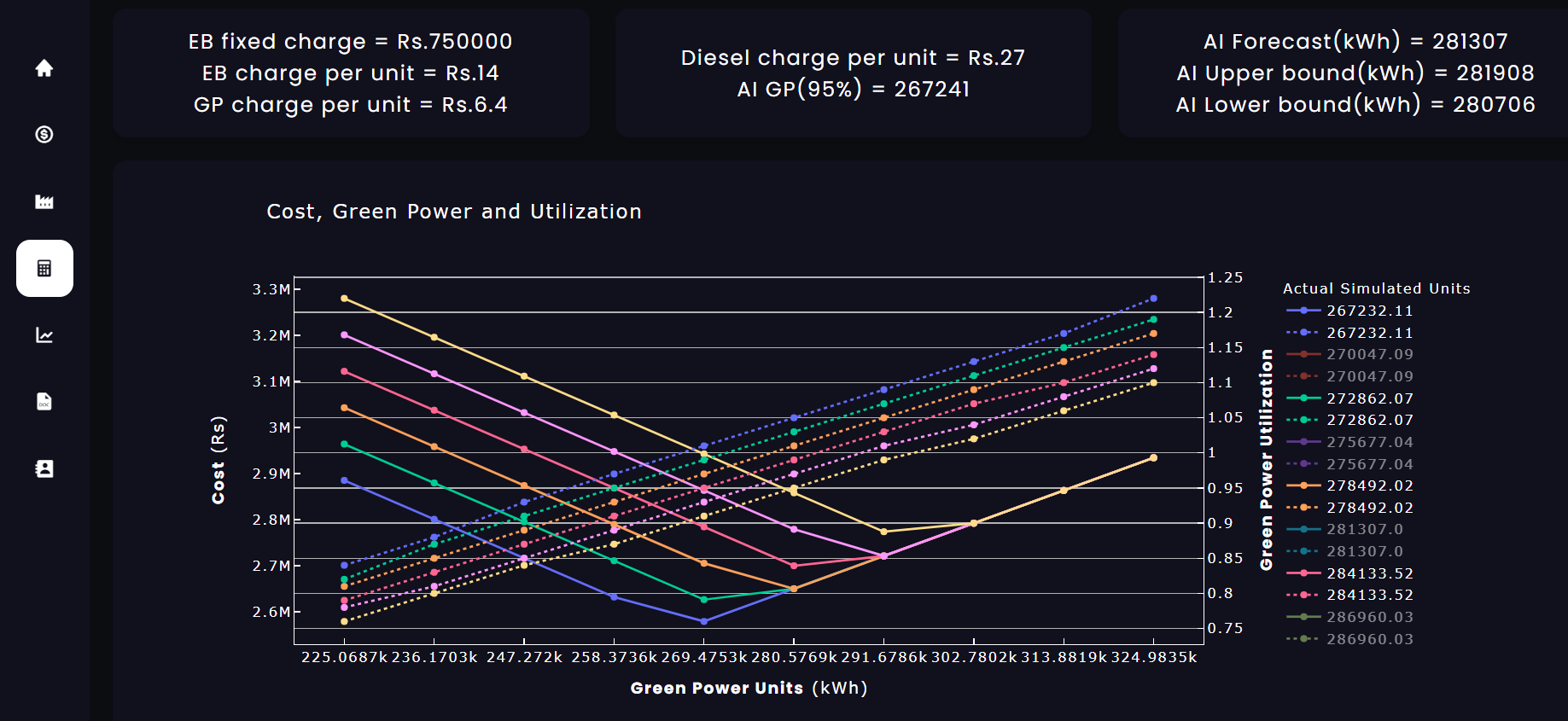} 
	\caption{Financial Modeling Tool}
	\label{FinancialModel}
\end{figure}

Green power vendors expect advance order placement and the client pays for energy units even if the units are underutilized. Hence there is a strong need for having an analytical tool to generate recommendation on green energy procurement in advance. Towards this end, we provide a financial modeling interface (refer Figure \ref{FinancialModel}) to recommend the green energy units by reducing overall cost while increasing the green energy utilization. Financial Modeling tool allows the user to conduct a variety of what-if analysis by simulating various combinations of green and conventional energy mix.
The figure shows simulated end of the month invoice amount for varying green power demand. Users can make decisions by arriving at a point where the cost variation is minimal and also the green power ratio is the maximum.

\subsection{Carbon Offset Recommendation Interface}
\begin{figure}[h]
	\centering
	\includegraphics[width=1\columnwidth]{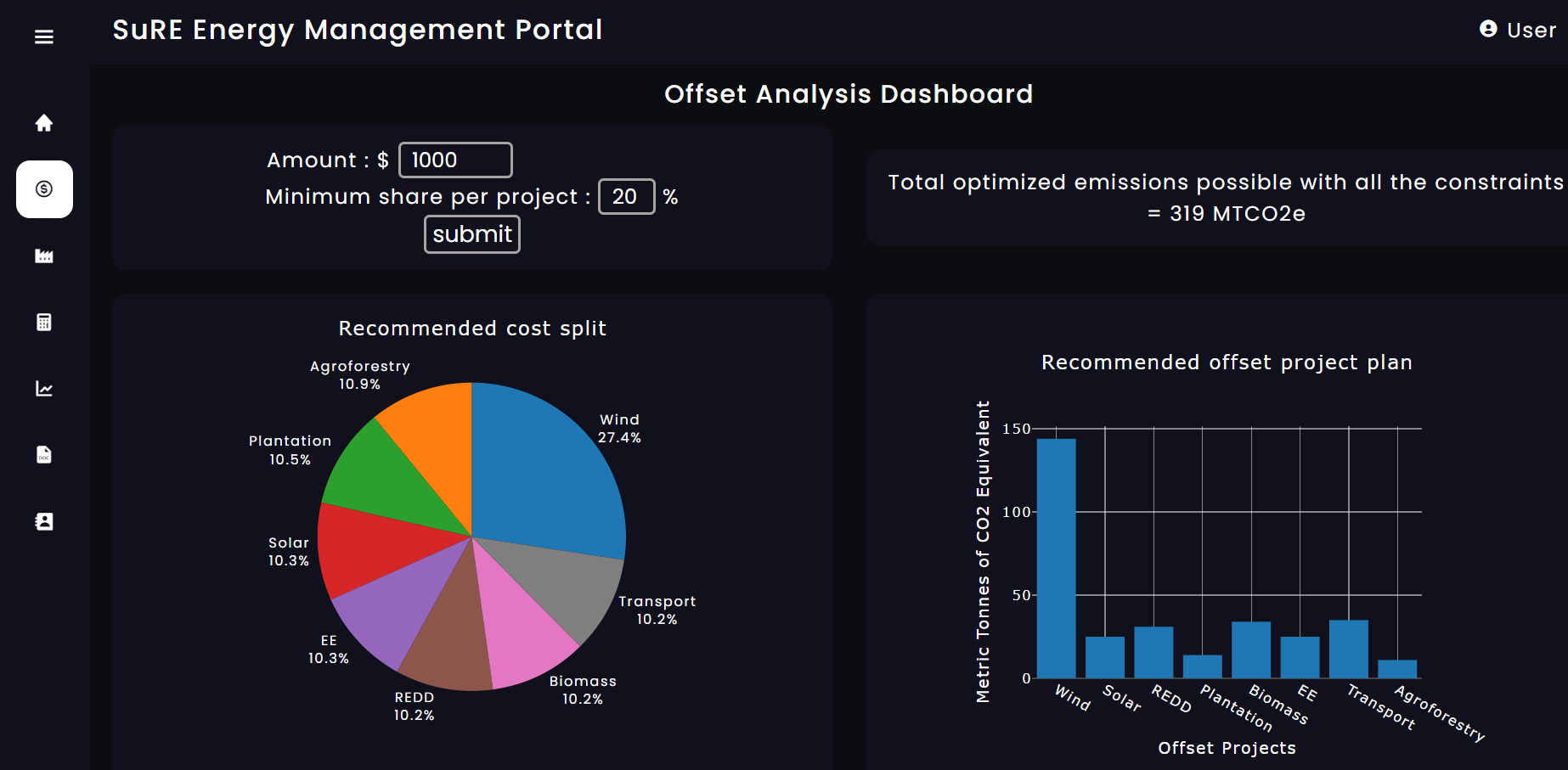} 
	\caption{Carbon Offset Recommendation Interface}
	\label{Recommendation}
\end{figure}

This view  (refer Figure \ref{Recommendation}) provides an efficient carbon offset plan to achieve Net-Zero emissions target. It gives two approaches to conduct the planning - (a) providing target investment amount (b) providing target offset in Metric Tonnes of CO2 equivalent (MTCO2e). Based on the selected options, recommendations will be provided on investments required in different offset projects. The tool also is configurable to select from a specific list of offset projects.

The deployed application is an end to end solution that provides users with seamless interactive experience with features comprising data upload, data ingestion automated training and evaluation of models on cloud in real time. 
This application also provides utilities to (a) analyze actual historic consumption with respect to the AI models based forecasts and (b) perform financial modeling to help in energy procurement and optimized carbon offset recommendation to nullify carbon emissions of facilities. 

\section{Results}
\label{results}

\begin{table*}[h]
	\centering
	\caption{Benefits Achieved and Potential Benefits of SuRE application adoption in terms of (a) Green Power (GP) Utilization in \% (b) Reduction in CO2 emissions (MTCO2e) and (c) Savings per unit. \\
	}	

	\begin{tabular}{|p{0.065\linewidth}|p{0.07\linewidth}|p{0.08\linewidth}|p{0.075\linewidth}|p{0.075\linewidth}|p{0.08\linewidth}|p{0.075\linewidth}|p{0.075\linewidth}|p{0.08\linewidth}|p{0.075\linewidth}|}
		\hline
		Facilities & \multicolumn{3}{|c|}{Before Adoption of SuRE} & \multicolumn{3}{|c|}{After Adoption of SuRE} & \multicolumn{3}{|c|}{Full Potential of SuRE} \\
		\hline
		& GP Utilization (\%) & Reduction in CO2 (MTCO2e) & Savings per unit (\$) & GP Utilization (\%) & Reduction in CO2$^\star$ (MTCO2e) & Savings per unit (\$)& GP Utilization (\%) & Reduction in CO2$^\star$ (MTCO2e) & Savings per unit (\$) \\
		\hline		       
		A & 85.79\% & 3349 & 0.092 & 90.67\% & 2261 & 0.098 & 96.05\% & 2396 & 0.104 \\
		\hline 
		B & 75.82\% & 1787 & 0.081  & 85.22\% & 1453 & 0.091  & 96.08\% &  1641 & 0.103  \\ 
		\hline
		C & 74.64\% &  2419 & 0.081 & 83.69\%  &  1609 & 0.090 &  96.85\% & 1864 & 0.104 \\
		\hline
		D & 87.41\% &  4451 & 0.095 & 91.39\% &  2147 & 0.099 & 95.55\% & 2243 & 0.103\\
		\hline
		Summary & 80.91\% &  12006 & 0.0872 & 87.74\% &  7470 & 0.0945 & 96.13\% &  8144 & 0.1035 \\
		\hline 
	\end{tabular}
	\\$^\star$Because of Pandemic situation there was a reduction in an overall electricity demand. This resulted in reduced CO2 emission savings.
	\label{all_results}
\end{table*}
\begin{figure*}[h]
	\centering
	\includegraphics[width=1.5\columnwidth]{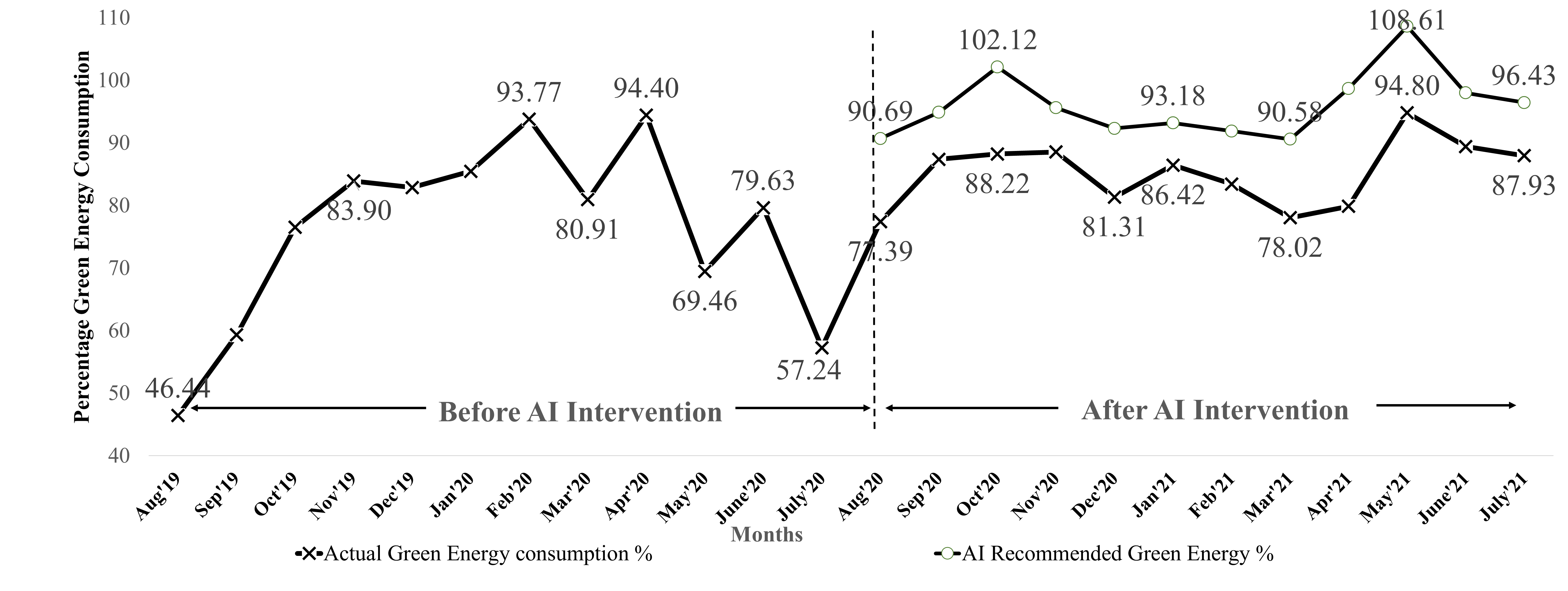}
	\caption{Green Power Consumption for Facility B before and after SuRE application intervention}
	\label{Results}
\end{figure*}
The facility teams were provided with the SuRE application to improve their Green Power (GP) utilization towards achieving Net-Zero emissions by 2023. 
The SuRE application is currently being utilized for four facilities (Facility A, B, C and D) from Aug 2020. The results presented in this section are for one year duration from Aug 2020 through July 2021. 
Before the adoption of SuRE application, manual techniques were used to estimate the monthly electricity demand for any facility. The facility teams further used heuristic approaches for carbon offset planning. 
Post this intervention, the facility teams have been using the SuRE application to forecast energy demand and determine monthly Green Power utilization. 
This enables facility teams to place accurate estimates for green power procurement.

In Table \ref{all_results}, we present comparison of before and after intervention of SuRE platform for the four facilities. It also presents the full potential benefits that can be achieved by using SuRE application. 
The key performance indicators used here are: Green Power utilization in \%, Reduction in CO2 emissions (MTCO2e) and Savings per unit in USD. 
\begin{itemize}
	\item Before the intervention of SuRE application, the average green power utilization for these facilities was 80.91\%. By considering the predictions from our solution, the potential to increase the green power utilization is 96.13\%. In our current deployment of SuRE application for 12 months, the facility teams were able to see an actual increase in green power utilization to 87.74\%. 
	
	\item The Reduction in CO2 emission due to green power utilization after adoption and with full potential is seen to be lesser as compared to before SuRE intervention. This is due to overall decrease in electricity demand due to the pandemic situation.  
	
	\item We also noticed that the overall cost savings per unit is increasing after the adoption of SuRE application, an increase of 8.5\%. This metric is normalized for total units, hence irrespective of the impact of pandemic we can see an increase in cost savings. The gap between potential and actual (After Adoption) increase in GP utilization depicts the adoption challenges of AI technology. 
	
	\item This 7\% improvement in green power share through AI based intervention has enabled the organization to reduce an additional carbon emissions equivalent to 559 tons of CO2 emissions and leading to cost savings of around 79K USD. By completely relying on AI recommendations the potential increase is 15\%. With this we could have achieved carbon reduction equivalent to 1303 tons of CO2 emissions and cost saving of 171K USD.
\end{itemize}

For Facility B, significant improvement in increase in green power utilization from 75.82 \% to 85.22 \% after SuRE adoption has been achieved. 
Figure \ref{Results} provides monthly utilization of green power before, after and potential impact of SuRE application for Facility B. For the time period \emph{Before AI Intervention}, drastic fluctuations in green power consumption has been observed. 
This is attributed to manual forecasting errors of overall demand and hence being conservative in the procurement of green power. 
For the time period \emph{After AI Intervention}, we can see the green power consumption trend showing less variance across the months. This has built a good trust with the facility leadership teams with commitment to adopt it with full reliance and scale-up has been now planned to multiple other locations globally.

By increasing the green power share the organization is now able to achieve multi-fold benefits:
\begin{enumerate}
\item Make progress towards Sustainability Initiatives of the organization, which has target of 100\% Renewable Electricity by 2023. 
\item Reduce Carbon Footprint by increasing share of Renewable Electricity for the selected facilities.
\item Achieve huge cost savings as green power is relatively cheaper than conventional resources in India.
\end{enumerate}

Towards this end, to fully achieve goals of Net-Zero emissions, our solution is being leveraged by the organization for carbon offset recommendations such as - procurement of renewable electricity certificates, investment in green power generation projects and agro-forestry projects.  

\subsection{Adoption Road map}
In this section we describe a broad sequence of steps any organization can follow to implement the SuRE model.

\begin{itemize}
	\item Based on the industry type and geography, identify the options which will help in achieving net-zero emissions like procurement of green power, generation of green power, investment in renewable electricity certificates, investment in other social projects which can provide the green credits. 
	
	\item Collect the available data sources or features which can provide an accurate forecast of energy demand, the data granularity can be hourly, daily or monthly.
	
	\item Utilize the modelling approach to forecast the occupancy of the building or any other features which may influence the energy demand within the facility.
	
	\item Forecast the energy demand using the independent variables (including the forecasted occupancy). 
	
	\item Generate recommendations for green power procurement and/or generation to achieve maximum utilization of the green power in the building operations.  
	
	\item Finally, to offset for the remaining energy and to achieve Netzero emissions, recommend other emission offset mechanisms like renewable electricity certificates, investments in agro-forestry etc. in a cost-effective way. 
\end{itemize}

\section{Conclusion}
\label{conclusion}
In this paper, we have showcased Sustainability using Renewable Electricity (SuRE) Application and its deployment for four large facilities of an organization for the period of one year. 
The SuRE application provides AI based models for energy demand forecasting which in turn is used for optimal recommendation for procurement of green power and help in creating a plan for achieving carbon emission reduction. 

Since the deployment of SuRE application for these facilities, the application has resulted in a significant increase in their overall green power utilization. 
This has resulted in significant cost savings along with reduction in carbon emission. 
To the best of our knowledge, this is one-of-a-kind applications for sustainable energy management that provides a unique amalgamation of various features, occupancy forecasting, energy demand forecasting and energy offset recommendations. 
Albeit the results were presented for the deployment in a specific geographic location, this solution is highly scalable and would be able to cater to any facilities across geographies where there is a need to build sustainable energy management operations.

As a subsequent work, we will continue to add more facilities in our application to create greater impact towards Global Net-Zero emission target. We also plan to extend the applicability of our work using advanced research areas like federated learning, where multiple facilities across different geographies and organizations can contribute in building accurate models in a federated manner. 

\textbf{Disclaimer:}
This content is provided for general information purposes and is not intended to be used in place of consultation with our professional advisors. \emph{SuRE} is the property of Accenture and its affiliates and Accenture be the holder of the copyright or any intellectual property over \emph{SuRE}. No part of this \emph{SuRE} may be reproduced in any manner without the written permission of Accenture. Opinions expressed herein are subject to change without notice.

\bibliographystyle{IEEEbib}
\bibliography{Sure_Ref}
\end{document}